\documentclass[letterpaper]{article}
\usepackage{aaai18}
\usepackage{url}
\usepackage{fullpage}
\usepackage{graphicx,curves}
\usepackage{amsmath}
\usepackage{amssymb}
\usepackage{cmbright}
\usepackage{latexsym}
\usepackage{multirow}
\usepackage[algoruled,vlined,linesnumbered]{algorithm2e}
\usepackage{wrapfig}
\usepackage{color}
\usepackage[table,xcdraw]{xcolor}
\newtheorem{conj}{Condition}
\usepackage{float}

\newtheorem{df}{Definition}

\newtheorem{theorem}{Theorem}

\newcommand{\bt}{\begin{theorem}\em}
	\newcommand{\et}{\end{theorem}}

\newcommand{\bea}{\begin{eqnarray}}
\newcommand{\eea}{\end{eqnarray}}
\newcommand{\bdf}{\begin{df}\em}
	\newcommand{\edf}{\end{df}}
\newcommand{\ben}{\begin{enumerate}}
	\newcommand{\een}{\end{enumerate}}

\numberwithin{equation}{section}
\numberwithin{theorem}{section}
\numberwithin{lemma}{section}
\numberwithin{df}{section}

\title{Evolving Real-Time Heuristics Search Algorithms with Building Blocks}

\nocopyright

\author{Md Solimul Chowdhury and Victor Silva \\
	Department of Computing Science \\ University of Alberta \\
	Edmonton, Alberta, T6G 2E8, Canada \\
	{\tt $\{$mdsolimu,vsilva$\}$@ualberta.ca}}

\begin{document}
	
\maketitle
	
\begin{abstract}
	The research area of real-time heuristics search has produced quite many algorithms. In the landscape of real-time heuristics search research, it is not rare to find that an algorithm $X$ that appears to perform better than algorithm $Y$ on a group of problems, performed worse than $Y$ for another group of problems. 
	If these published algorithms are combined to generate a more powerful space of algorithms, then that novel space of algorithms may solve a wide distribution of problems efficiently. Based on this intuition, a recent work \cite{bulitko2016evolving} has defined the task of finding a combination of heuristics search algorithms as a survival task. In this evolutionary approach, a space of algorithms is defined over a set of building blocks (published algorithms) and a simulated evolution is used to recombine these building blocks to find out the best algorithm from that space of algorithms. In this project, we extend the set of building blocks by adding one published algorithm, namely - lookahead based $A^{*}$ shaped local search space generation method from $LSS\!-\!LRTA^{*}$, plus an unpublished novel strategy to generate local search space with Greedy Best First Search. Then we perform experiments in the new space of algorithms, which shows that the best algorithms selected by the evolutionary process have the following property: the deeper is the lookahead depth of an algorithm, the lower is its suboptimality and scrubbing complexity.  
\end{abstract}
	
\section{Introduction}

Real-time Agent-Centric search is an important area of Artificial Intelligence, where plan search and plan execution happens in real-time in an interleaving fashion. In a real-time setting, an agent executing a real-time heuristics search algorithm, has access only to the limited vicinity of the current state. A real-time heuristics search algorithm can be described as an iteration of the following three steps: 
\begin{itemize}
	\item [i.] search in the immediate neighborhood of the current state to select the best state to move, 
	\item[ii.] update the heuristic value of current state by examining the heuristics value of its neighboring states, and then
	\item[iii.] from the current state, move to the best neighboring state found in (i)
\end{itemize}
Starting with the cornerstone algorithm $LRTA^{*}$ \cite{Korf90}, the area of real-time search has produced a great number of algorithms, focusing on improving search \cite{sven09lssLRTA}, learning \cite{BulitkoL06} \cite{wbLRTA} and movement selection \cite{expendable} of the agent.  
		
Empirical results with great number of algorithms reveals one interesting point: performance of an real-time heuristics search algorithm is highly dependent on the problem. For example, \emph{LSS-LRTA*} \cite{sven09lssLRTA}, which uses \emph{A*} to generate a \textit{Local Search Space (LSS)}, was proposed as an improvement over basic \emph{LRTA*} algorithm that uses \emph{Breadth First Search}, for generating LSS. While \emph{LSS-LRTA*} generates better quality solution for \emph{Grid with Random Obstacle} benchmark, the basic \emph{LRTA*} generates better quality solution for the \emph{Maze} benchmark. 
		
		
In \cite{BulitkoL06}, a parametrized approach had been made to unify many of the published real time heuristic search algorithms into a single framework, named \emph{LRTS}. The idea was to extract the core techniques of these algorithms in LRTS. The empirical results show influence of the parameters on each other (Table 9 of \cite{BulitkoL06}), but this empirical results does not give a clear indication on what parameter (algorithm) to choose for a given problem. 
		
In a more recent work \cite{bulitko2016evolving}, the author has attempted to solve the problem of algorithm selection by applying evolutionary methods in a space of algorithms. The idea is the following: Given a space of algorithms $P$ as building blocks, and problem set $B$, apply $P$ algorithms on $B$ and then let an evolutionary process to select the best set of building blocks $P' \subset P$, by recombining the building blocks in each generation in a performance based way.

Building blocks in \cite{bulitko2016evolving} includes \emph{Depression Avoidance} \cite{HernandezB14}, \emph{Online pruning of dead states} \cite{expendable} and \emph{backtracking} \cite{ShueZ93}, and \emph{weighted learning} \cite{wbLRTA}. In this project, we extend this building blocks by adding two more algorithms:
\begin{itemize} 
	\item [i.] A* based LSS generation method of LSS-LRTA* algorithm and 
	\item [ii.] As mentioned earlier in this section, \emph{LSS-LRTA*} equipped with A* algorithm to generate LSS, performs worse than the basic LRTA* algorithm, that uses BFS for the same purpose. One interesting question is, how greedy Best First Search (BSFS) (that takes only $h$-value into account to determine what state in the LSS to expand next) to generate LSS, performs compared to the other two methods. So, here we propose another strategy for generating LSS with greedy BSFS and include it in our extended list of building blocks as the second enhancement.   
\end{itemize}
The rest of this paper is organized as follows: We present the problem formulation in the next section, which is followed by the related work. In section 4, we present the details of our approach, including the devised algorithm and implementation. Theoretical analysis of the devised algorithm is presented in section 5. Section 6 presents empirical evaluation of the algorithm presented in section 4. In section 7, we present a discussion on this empirical results. In the last section, we conclude our paper and discuss about some future directions of this work.
\section{Problem Formulation}

The search problem \textbf{S} can be defined as the tuple $(S,E,c,s_{0},s_{g},h)$ : given a graph $G(S,E)$, where $S$ is a set of states and $E$ is a set of edges between any two distinct nodes $s \in S$. We assume that the graph $G$ is connected, stationary and undirected. Traveling through each edge $e$ incurs a cost $c>0$. 
		
Given a search problem \textbf{S}, a search agent operates in the search graph $G \in \textbf{S}$, starts at the starting node $s_0$, travels through the states space of $G$ to reach the goal node $s_{g}$. While traveling to $s_g$ from $s_0$, the agent incurs a \emph{solution cost}, which is the summation of cost of traveling through all the edges in its path to $s_g$ from $s_0$. By $N(s)$, we define a set of neighboring state of a state $s\in S$. 
		
During the search, the agent has access to a heuristic value $h$ that is an cost estimate from the current position of the agent to the goal state $s_{g}$. This heuristic is not assumed to be consistent or admissible. The agent uses this heuristic to search for the goal state $s_{g}$ and updates the heuristics value as required. However, the heuristics value at the goal state $s_g$ is immutable and $h(s_{g}) = 0$. For all other states $s \ in S$, at any time $t$, the heuristic is $h_{t}$.  The optimal heuristic is defined as $h^{*}(s_{0})$. A state $s \in S$ is also associated with a $g$ value, that denotes the distance from the start state $s_0$ to $s$, with $g(s_0) = 0$.  
		
To measure the performance of an agent, we use the suboptimality metric $\alpha$, obtained as a ratio of the solution cost incurred by the agent to the optimal path $h^{*}(s_{0})$. For our work, \textit{scrubbing complexity} is a also a relevant metric and it is represented by $\tau($\textbf{S}$)$. $\tau$ measures the average number of states visit by an algorithm while solving a search problem. The lower the values of $\alpha$ and $\tau$ are for a given algorithm, the better is the performance.  
		
The real-time heuristics search methods are \textit{agent centered} search methods, where an agent is required to search in the graph $G$, using local knowledge given a neighborhood bound. Besides, the search must be performed in real-time, meaning that the agent must commit an action, under a defined timespan.
		
We say that an agent (algorithm) is complete iff. it can solve any search problem in a finite timespan. While solving a given problem, to prevent an agent from running indefinitely, a bound $\alpha_{max}$ is used as the upper bound of the total travel cost. When the agent exceeds the $\alpha_{max}$ value, we assume that it is not capable of solving that problem. 
		
In this paper, we extend the building blocks of algorithms by adding two different building blocks. This extended building blocks defines a new space of algorithms. With this, in the following, we formulate the problem of our paper:
		
\begin{center}
	\textit{We assume a set of problems $B$ and the new space of algorithms $A$. Then by using the evolutionary approach of \cite{bulitko2016evolving}, we want to find out the algorithm from $A$ that is best at minimizing $\tau$ and $\alpha$.}
\end{center}
		
		
		
		
\section{Related Work}

In the research area of AI search, the performance based selection of algorithm is not entirely new. The idea is to use a set of algorithms under a hood and exploit the best of these algorithms to solve a particular search problem. Formally, this type of approach is known as portfolio based approach.
				
In SAT and automated planning research, portfolio based problem solving has already been pursued. As for SAT, authors of \cite{SATZilla}, have proposed a portfolio based approach, named \emph{SATzilla}, where a distribution of problem instances and a set of component solvers are taken as input to construct a portfolio that optimizes a given objective function, such as, mean runtime or number of instance solved. Given a set of SAT problems, \emph{SATzilla} extracts some features from that problem, trains a machine learning algorithm with these features. Then, when a new problem is given, the machine learning algorithm predicts the best component solver to solve that problem. ArvandHerd \cite{ArvandHerd} is a portfolio based parallel automated planner. It runs an instance of domain-specific deterministic planner named LAMA \cite{lama} and instances of random walk based planner named $Arvand$ \cite{arvand} at the same time, in different parallel cores. All the planners run in parallel, until a solution is found or until a pre-set maximum running time for all the planners expires. 
		
Both of \emph{SATzilla} and \emph{ArvandHerd} performed brilliantly in yearly held competitions organized by respective communities.
		
In another related work \cite{LPG}, the authors have studied automated parameter tuning in a highly parametrized stochastic local search based planner named $LPG$. $LPG$ has 62 parameters and over $6 \times 10^{17}$ configurations. To choose a combination from this huge space of configuration settings is an arduous task. The idea of automated tuning presented in this paper is this:  the planner LPG is augmented with a state-of-the-art automatic parameter tuning algorithm named $paramILS$. For some selected problem of some given domains, $paramILS$ runs experiments with various configuration settings from the valid parameter settings of $LPG$ and selects the best configuration setting based on \textit{mean runtime} of the configuration settings used to run the given problems. $LPG$ augmented with $paramILS$ outperforms the state-of-the-art planners.

To the best of our knowledge, these automatic algorithm selection techniques described above has not been applied to the area of real-time heuristics search yet. But, we think these techniques can be applied to the real-time heuristics search algorithms as well. In the following, we discuss about adoption of portfolio based approach in real-time heuristics search algorithm selection. 

Following the approach of SATzilla, a machine learning algorithm can be trained with a set of search problem features. The goal for the machine learning algorithm is to learn to predict best possible algorithm, given a new search problem. We are aware of some unpublished work along this line that purses deep learning to select the best possible algorithm. 

The approach of ArvandHerd should also be amenable to real-time heuristics search algorithms. The idea is to run several real-time heuristics search algorithms in parallel. There can be two modes of executions: a) continue with execution of the given problem with different algorithms in different parallel cores, until a solution is found by any of the algorithms. b) continue execution of the algorithms on the given problem, until the desired suboptimality is achieved by any of these algorithms.

The idea of parameterized LPG should also be applicable in Real-time heuristics search algorithms. Given a set of problems in each map (or a set of maps), a space of algorithms (combination of parameters) can be run in a parameter tuning system, such as paramILS, to experiment with different parameter settings. Based on the pre-set performance criteria, the parameter tuning system finds out the performance of these algorithms on the given problems set. 

To the best of our knowledge, the work of \cite{bulitko2016evolving} is the first of its kind in the area of real-time heuristics search. It presents a novel approach for performance based algorithm selection, where it represents some published algorithms as a set of building blocks. The idea is to use simulated evolution to recombine the building blocks in a performance based way. The preliminary results of this evolutionary approach are promising, which is one of the motivating factors of our paper. 
		
\section{Proposed Approach}

The idea of the building block approach is to create a space of algorithms, where an algorithm in that space corresponds to a specific combination of building block values. In our proposed approach, we use all the six building blocks from \cite{bulitko2016evolving}, plus two new building blocks from \cite{sven09lssLRTA}. In the following we describe each of these building blocks.	
\subsection{Building Blocks}
In \cite{bulitko2016evolving}, two groups of building blocks are used, namely- \textit{Movement Rule Building Blocks} and \textit{Learning Rule Building Blocks}. In addition to this two groups, we use another group of building blocks, namely - \textit{Local Search Space (LSS) generation Building Blocks}. 
\paragraph{Movement Rule Building Blocks}
\paragraph{Backtracking} The backtracking block performs a regression of the agent to a previous state, upon the detection of an underestimation in the initial heuristic estimation. The agent performs the heuristics update not only in the current state, but also in the previous ones. The backtracking building block is denoted as \texttt{backtrack}. 
		\paragraph{Expendable States}
		 The building block \texttt{expendable} is responsible for removing expendable states from the search space. Formally, expendable states are the ones, visit to which are not required to reach the goal state and thus expendable. This reduces the total size of the number of states to be visited, making the search space smaller.
		 \paragraph{Depression Avoidance}
		 Depression Avoidance ($d$) is a building block that detects the states in the neighborhood of the current state, whose heuristics values lead to depressions. For making the next move, this scheme considers only those states in the neighborhood, on which heuristics value raise does not exceeds a given threshold. When these states are detected, the agent avoids visiting these states, yielding less scrubbing.
		 
		\paragraph{Learning Rule Building Blocks}
		\paragraph{Heuristics Weighting} Heuristic weighting ($w$) is the block that performs an acceleration in the learning process. This block performs a weighting in the heuristic learning by using $w>1$ as a multiplicative factor. When the block uses this technique, the agent is less likely to re-visit states, preventing the scrubbing to happen and encouraging a smaller suboptimality.
		\paragraph{Learning Operator}
		Learning Operator block ($lop$) allows the agent to choose between four different basic operations during the heuristic learning process. Those operators are: $min, average, median$ or $ max$.
		\paragraph{Lateral Learning}
		The lateral learning building block let heuristics learning to be performed not just in the current state, but in part of the neighborhood state, including the current state. This neighborhood size is defined by the $beam width (b)$ parameter.
		
		\paragraph{LSS generation Building Blocks}
		\paragraph{Lookahead Depth}
		In the planning phase of the real-time heuristics search algorithms, states of the local neighborhood of the current state is examined to determine the best state to move next. While executing planning, an algorithm can lookahead a given depth in its neighborhood. A lookahead depth \texttt{lookahead} is an integer number that represents the number of states that are expanded during the planning phase. Intuitively, an algorithm looks deeper into the neighborhood to determine best state to move within the \texttt{lookahead} distance of the current state.    
		\paragraph{Lookahead Method}
		A lookahead in a search graph can be performed various ways. In \cite{sven09lssLRTA}, the authors have used $A^{*}$ search method to perform lookahead in the neighborhood of a given current state to generate the local search space. In our proposed approach, we introduce another building block, named \texttt{lookaheadMethod}, where lookaheadMethod can have two values, $A^{*}$ search method and $Greedy$ search method. When \texttt{lookaheadMethod} = $A^{*}$, $h+g$ value is used to obtain the next best state to expand and when \texttt{lookaheadMethod} = $Greedy$, $h$ value is used to obtain the best state to expand next.
		\subsection{Real-time Heuristic- Search Algorithm with Extended building blocks}
		
In this section, we present a real-time heuristics search algorithm with the extended building blocks. We adopted the pseudo-code presented in \cite{sven09lssLRTA} and have modified the pseudocode to accommodate the building blocks from our set of building blocks. 
		 
First, we present a high-level description of the algorithms, then we present all the technical details. Our real-time heuristic search algorithm with building block takes a search problem as input, starts with the start state and reaches the goal state by iteratively executing the four operations in an interleaving fashion: i) Determination of local search space, ii) Determination of the best frontier state in the local search space to perform the next move, iii) Update of the $h$-value of the local search space and iv) Travel to best frontier state. For i), starting from the given start state, a search is performed iteratively by using either $A^{*}$ or $Greedy$ method (determined by the \texttt{lookaheadMethod} building block), until it expands number of nodes equal to the provided lookahead depth (determined by \texttt{lookahead} building block.). In ii), the best frontier state (that has the minimum of $h+g$ value) of the local search space is obtained and it becomes the next state to move. In iii), $h$ value of the local search space determined by i) are updated. Essentially it is Dijkstra procedure that updates the $h$-values of the local search space. In iv), the agent travels from the current start state to the best state in the local search space (determined by ii), by using a tree pointer prepared in (i). Then, that best state becomes the current state and the search continues from there.    
		 	
In Algorithm 1 to Algorithm 4, we present the pseudo-codes of our approach.   The main procedure is shown in Algorithm \ref{buildingBlockAlg}. This algorithm accepts a search problem $(S,E,c,s_0,s_g,h)$ and a set of control parameters $w,b, lop, da$, \texttt{lookahead, lookaheadMethod, expendable} and \texttt{backtrack} as input. It assigns the start state $s_0$ to the current state $s_t$ (Line 2). Then it employs a while loop, which executes until the goal state $s_g$ is reached (Line 3 to 9). In each iteration of the while loop, it first generate a local search space by calling the $generateLSS$ procedure, that returns the generated local search space inside the Closed and Open list (Line 4). The closed contains the states expanded by $generateLSS$ and the Open list contains the frontier states yet to be expanded by $generateLSS$. After that, from the Open list, it obtains a state $s'_{goal}$, such that $s'_{goal}$ has the minimum of $h+g$ value among all other states in the Open list (line 5). $s'_{goal}$ is best state in the local search space to which the algorithm moves next. So, $s'_{goal}$ is the next current state. After determining $g'_{goal}$, it updates the $h$-values of the states in the local search space, by calling the $updateLSS$ procedure (line 6). After the update, it executes the actual movement from $s_t$ to $s'_{goal}$ by calling the $moveToBestFrontier$ (line 7) procedure. Lastly, it updates the current state $s_t$ by assigning $s_{goal}$ to $s_t$ (line 8).  
		 
Algorithm \ref{buildingBlockAlg} receives eight building blocks, but, currently our algorithms uses five of these building blocks, namely: $w,da,lop$, \texttt{lookahead} and \texttt{lookaheadMethod}.

\begin{algorithm}
	\DontPrintSemicolon
	\label{buildingBlockAlg}
	\caption{Real-time Heuristics Search w/Building Blocks}
	\KwIn{Search problem ($S,E,c,s_0,s_g,h$), control parameters $w,b, lop, da$, \texttt{lookahead, lookaheadMethod, expendable, backtrack}}
	\KwOut{path ($s_0, s_1, \dots s_T$), $s_T = s_g$}
	$t \leftarrow 0$\;
	$s_t \leftarrow s_0$\;
	\While{$s_t \ne s_g$}{
		%
		$[Open; Closed] \leftarrow generateLSS(s_t,w, lop, da$, \texttt{lookahead, lookaheadMethod})\; 
		$s'_{goal}  \leftarrow arg min_{s'\in Open} h(s') + g(s')$ \;
		$updateLSS(Open, Closed,h,w, lop, da$, \texttt{lookahead, lookaheadMethod})\;	
		$moveToBestFrontier(s_t,s'_{goal})$\;
		$s_{t+1} \leftarrow s'_{goal}$ \;
		$t \leftarrow t+1$ \; 
	}
	$T \leftarrow t$
\end{algorithm}

The procedure $generateLSS$ (Algorithm 2) takes five parameters: $s_t, da$, \texttt{lookahead} and \texttt{lookaheadMethod}. First, it initializes $g$, $Open$ and $Closed$ list (line 1-6). The $Closed$ and $Open$ list to store states of the expanded states of thee LSS and frontier states of LSS respectively. For each of the states $s \in S$, it assigns $\infty$ to $g(s)$, except the $g(s_t)$, which is initialized with 0.  Then it employs a while loop until a stopping condition is met, that is, the flag $continueSearch$ turns to be false. $continueSearch$ is set to true (line 27) inside the loop body if a) $g(s_g)$ is greater than minimum $f=h+g$-value state in the $Open$ list and b) lookahead \textit{expansions} is yet to reach the \texttt{lookahead}. In each iteration of the while loop, it first applies i) depression avoidance (line 12), i.e., if $da$ is true, it takes only that states from the immediate neighbourhood whose $h$-value has not raised by a threshold $th$. ii) Otherwise, It considers the whole neighbourhood. The states obtained from (i) or (ii) then put into the list $mOpen$. Then, it proceeds with determining which node from $mOpen$ is to be expanded next. If \texttt{lookaheadMethod} is $Greedy$ then it selects the state $s$ that has the least $h$-value in the $mOpen$ list, otherwise if \texttt{lookaheadMethod} is $A^{*}$, it selects the state $s$ that has the least $h+g$ value in the $mOpen$ list. $s$ is then deleted from $mOpen$ and added to the $Closed$ list (line 18,19). Then, for each valid action $a$ at state $s$, it then determines the successor state $s_{succ}$ of $s$ (line 20 to 26). The state $s_{succ}$ is pushed into to the frontier (to be a new search frontier), if its $g$ value is greater than $g$+ edge cost of travelling (c) from $s$ to $s_{succ}$. In this case, $g(s_{succ})$ is  updated by $g(s)+c$.  A \textit{tree} pointer is used to track the edges between $s$ and new frontier nodes $s_{succ}$. Note that this \textit{tree} pointer is used by the $moveToBestFrontier$ procedure to travel through the optimal path to reach the next goal.
\begin{algorithm}
		
	\DontPrintSemicolon
	\label{generateLSS}
	\caption{Generation of the Local Search Space}
	\KwIn{$s_{t}, s_{g}, h_{0},h, w, da$, \texttt{lookahead, lookaheadMethod}}
	\KwOut{$[Open, Closed]$}
	\ForEach{$s \in S$}{
		$g(s) \leftarrow \infty$ \;
	}
	$g(s_{t}) \leftarrow 0$ \;
	$Open \leftarrow \emptyset$\;
	$Closed \leftarrow \emptyset$ \;
	$Open \leftarrow Open \cup \{s_{t}\}$ \;
	$expansions \leftarrow 0$ \;	
	$continueSearch \leftarrow true$ \;
	$mOpen \leftarrow Open$ \;
	\While{$continueSearch$}
	{
		$expansions \leftarrow expansions + 1$\; 
		\If {$da$}{
			$mOpen \leftarrow \{s' \vert s' \in \! mOpen \; AND \; h(s') - h0(s') < th \}$\;
		}

		\eIf{$\texttt{lookaheadMethod}="Greedy"$}{		
			$s \leftarrow arg\; min_{s'\in mOpen} h(s')$\;
		}
		{
			$s \leftarrow arg\; min_{s'\in mOpen} h(s') + g(s')$\;
		}
		$mOpen \leftarrow mOpen \setminus \{s\}$ \;
		$Closed \leftarrow Closed \cup \{s\}$ \;
		\ForEach{$a \in A(s)$}
		{
			$s_{succ} \leftarrow Succ(s,a) $ \;
			\If{$g(s_{succ}) > g(s) + c(s,a)$}
			{
				$g(s_{succ}) \leftarrow g(s) + c(s,a)$ \;
				$tree(s_{succ}) \leftarrow s$ \;
				\If{$s_{succ} \notin mOpen$}
				{
					$mOpen \leftarrow mOpen \cup \{s_{succ}\}$ \;
				}	
			}				
		}
		$continueSearch \leftarrow$ $g(s_{g}) > min_{s'\in mOpen} h(s') + g(s')$ AND $expansions < \texttt{lookahead}$ \;
	}
		
	Return 	$[mOpen, Closed]$ \; 
\end{algorithm}

\begin{algorithm}
	\DontPrintSemicolon
	\label{updateH}
	\caption{Update Heuristics}
	\KwIn{$h, Open, Closed,w, lop, da$, \texttt{lookahead, lookaheadMethod}}
	\KwOut{updated $h$}
	\ForEach{$s \in Closed$}{
		$h(s) \leftarrow \inf$ \;	
	}
	\While{$Closed \ne \phi$}{
		$s = arglop_{s'\in Open} h(s')$\;
		$Closed = Closed - \{s\}$ \;
		$Successors = \{s' \vert Succ(s',a) = s\}$ \;
		\ForEach{$s' \in Successors$}{
			\If{$s' \in Closed$ \&\& $h(s') > c(s',a) + h(s)$}{
				$h(s') \leftarrow w*(c(s',a) + h(s))$ \;
				\If{$s' \notin Open$}{
					$Open \leftarrow Open \cup \{s'\}$\;			
				}
			}	
		}	
	}
\end{algorithm}

The $updateHeuristics$ (Algorithm 3) employs the Dijkstra style update of $h$-values of the LSS. It iteratively updates the $h$-values of the already expanded states (in the $Closed$ list), which are the neighbors of the best frontier state $s$ (in the $Open$ list) determined by the learning operator $lop$. While updating the heuristics, the update takes the building block $w$ into account.

After updating the heuristics for the local neighborhood by using $updateHeuristics$ function, the $moveToBestFrontier$ procedure (Algorithm 4) is called from Algorithm 1. It adds the travel cost of travelling from the current state $s_t$ to the next goal state $s_{g}$ (determined by Line 5, Algorithm 1) with the $distanceTraveled$ (total distance traveled) to account for total distance traveled so far. It uses the $tree$ pointer to determine the path from $s_t$ to $s_g$. During the tree pointer update in Algorithm 2, the pointer from the successor node $s'_t$ of the current state $s_t$ to $s_t$ can be overwritten. As this procedure uses a while loop that executes until it enters $s_t$ along the optimal path (line 11 to 15), Algorithm 4 ensures that a pointer from a neighboring state of $s_t$ to $s_t$ exists in the tree (line 1 to 7). Note that the procedure $moveToBestFrontier$ was not explicitly defined in \cite{sven09lssLRTA}. In our paper, we have defined it explicitly.
\begin{algorithm}
	\DontPrintSemicolon
	\label{moveTo}
	\caption{Move to the best frontier node}
	\KwIn{$s_t,s_g$}
	\KwOut{$none$}
	\If{$s_t \notin tree$}{
		\ForEach{$a \in A(s_t)$}{
			$succ_{s_t} \leftarrow succ_{s_t} \cup succ(s_t,a)$\;
		}
		\ForEach{$s \in succ_{s_t}$}{
			\If{$s \in tree$}{
				$tree(s) = s_t$ \;
				\textit{break}
			}
		}
		
	}
	$s \leftarrow s_g$ \;
		
	$prevState \leftarrow tree(s)$ \;
	$a \leftarrow \{a' \vert succ(prevState,a') = s\}$ \;
	\While{$prevState \neq s_t$}{
		$distanceTraveled \leftarrow distanceTraveled + c(prevState,a)$ \; 
		$s \leftarrow prevState$ \; 
		$prevState \leftarrow tree(s)$ \; 
		$a \leftarrow \{a' \vert succ(prevState,a') = s\}$ \; 
	}
\end{algorithm}

\subsection{Implementation}
We have implemented Algorithm 1 to Algorithm 4 in MATLAB. In our implementation, we have worked on top of the code-base provided by the course instructor.

In our implementation, the following files from that code base are modified:
\begin{itemize}
	\item generateSupportDBPerMap.m : We have extended the geneMax and geneMin to included maximum and minimum lookahead depth.  
	\item uLRTA.m : This file contains the matlab function named uLRTA, which implements the real-time heuristics search algorithm with the building blocks of \cite{bulitko2016evolving}. We have implemented Algorithm 1 by modifying the function uLRTA. 
\end{itemize}
To implement Algorithm 2, we have created a separate MATLAB function named $generateLSS$. Algorithm 3 and Algorithm 4 are implemented within the uLRTA function. 

In \cite{sven09lssLRTA}, it does not present any explicit algorithm for movement through the optimal path. In this report, we have this explicit in Algorithm 4. While implementing Algorithm 4, we have faced two issues, that we needed to overcome. In the following, we are discussing those two problems.

First, Algorithm 2 generates the local search space by looking-ahead into the local neighbourhood. Additionally, it identifies an optimal path to the next goal state $s'_{goal}$ from the current state $s_t$. A \textit{tree} is used to store the pointers from a node to its parent node along that optimal path. Precisely, a \textit{parent} node is inserted into the tree in it's \textit{child} position (in the optimal path) via the following assignment: \textit{tree(child) = parent}. During the implementation, we observed that in quite a few occasions the current state $s_t$ goes missing from from tree. This is problematic, because Algorithm 4 does not terminate if the current state $s_t$ is not in the \textit{tree}. 
	      	
Our investigation reveals the following fact that explains how $s_t$ goes missing from the $tree$: At $i^{th}$ execution of the while loop of the current invocation to Algorithm 2, $tree(s_{succ})$ contains $s_t$ (where $s_{succ}$ is a neighbor of $s_t$). But, at $j^{th}$ execution of the while loop ($j>i$) of the current invocation of Algorithm 2, $tree(s_{succ})$ is overwritten by a neighboring state of $s_{succ}$.  While executing movement through the optimal path, in our implementation, we fixed this problem by inserting $s_t$ into the tree if it found to be missing. This is reflected in Algorithm 4.

Secondly, we have observed another scenario during our implementation of Algorithm 4. Algorithm 4 uses a while loop to construct the path starting from the goal state $s'_{goal}$ to the current start state $s_t$. Inside the while loop, it gets the parent state of a \textit{state} by using $tree(state)$. In some rare occasions, this extraction of the parent state of a state leads to a cycle. For example, starting with state $x$, in a iteration of the while loop of the Algorithm 4, it produces $y=tree(x)$, then in the next iteration it produces $z=tree(y)$ and in the next iteration of the while loop it produces $x=tree(z)$ and the procedure enters into a cycle. 	In case such cycle appears, we choose a strategy to terminate the movement execution.

We debugged our code for possible bugs that could lead to such cycle, but found that the cycle is appearing \textit{naturally}. This scenario requires further investigation. 

\section{Theoretical Analysis}

We have extended the building blocks of algorithms defined in \cite{bulitko2016evolving} by introducing two building blocks, namely- \texttt{lookahed} and \text{lookaheadMethod}. 

Two questions arises with these extension: 1) Does the Algorithm \ref{buildingBlockAlg} terminates in finite time with the extended building blocks? 2) Does the Algorithm \ref{buildingBlockAlg} always reaches the goal? In the following, we present answers to these questions.

\subsection{Termination of Algorithm \ref{buildingBlockAlg}}

Starting with the start state, Algorithm \ref{buildingBlockAlg} loops through LSS generation, heuristics update of the LSS and movement to the best LSS frontier, until the goal is reached. In each iteration, it selects the best frontier node (from Open) of the LSS to make a move towards the goal state. In case a state $s$ is revisited, Algorithm 3 updates the $h$-values of it's neighborhood states. These updates let the agent steps out of the neighborhood. Thus, the search always progresses toward the goal. An exception occurs when the Open becomes empty and the agent does not have any available move. In this case, the problem is not solvable and the search terminates.  

To claim that Algorithm  \ref{buildingBlockAlg} terminates in finite time, we need to show that the procedures $generateLSS$, $updateHeuristics$ and $moveToBestFrontier$ also terminates in finite time. The main loop of of the procedure $generateLSS$ terminates after the $expansions$ reaches the provided lookahead depth \texttt{lookahead}. So, the termination of the procedure $generateLSS$ of Algorithm \ref{generateLSS} is guaranteed. For the $updateHeuristics$ procedure of Algorithm \ref{updateH}, the $Closed$ list contains already expanded nodes by the latest call to the procedure of Algorithm \ref{generateLSS} and thus $Closed$ is finite in size. Termination of Algorithm \ref{updateH} is guaranteed by the finite size of the $Closed$ list. The procedure in $moveToBestFrontier$ of Algorithm \ref{moveTo} also terminates in finite time, as it traverses backward from the next goal state to the current state with the guidance of the \textit{tree} pointer. Thus, total number of iterations of the while loop of Algorithm \ref{moveTo} is bounded by the length of the path (identified by the \textit{generateLSS} procedure) starting from the current state to the next goal state. 

Thus, Algorithm \ref{buildingBlockAlg} terminates in finite time. 

\subsection{Goal Reachability of Algorithm \ref{buildingBlockAlg}}
In this section, we derive a necessary condition for goal reachability of Algorithm \ref{buildingBlockAlg}. 

Algorithm \ref{buildingBlockAlg} uses generateLSS method to obtain the best search frontier state (from the Open list) of the LSS. If generateLSS keeps the Open list non-empty, Algorithm 1 always has a move. For any state $s$ that is to be expanded next by generateLSS, the expansion of $s$ adds any $s_{succ} \in N(S)$ into the Open list, iff. $g(s_{succ}) > g(s) + cost(s,s_{succ})$ is true. Now, at any stage of execution of generateLSS, prior to the expansion, if Open list contains only one state $s'$ and for all of $s'_{succ} \in N(s')$ if $g(s'_{succ}) <= g(s') + cost(s,s'_{succ})$, then Open will become empty after $s'$ is expanded. And in this case, algorithm 1 will not have any moves left. So, we derive the necessary condition for goal reachability of Algorithm \ref{buildingBlockAlg}:
\begin{conj}[Goal Reachability Condition]
	Let $Open^i$ denotes the set of states in the Open list at the $i^{th}$ execution of the while loop of the generateLSS procedure. Algorithm \ref{buildingBlockAlg} will reach goal, if with $Open^{i} = \{s\}$, there exists a $s_{n} \in N(s)$ for which $g^{i-1} (s_n) > g^{i-1}(s) + cost(s,s_n)$.
\end{conj} 
If the above condition holds, then Open  list will remain non-empty and the Algorithm \ref{buildingBlockAlg} always has a move. Then by guaranteed termination of Algorithm \ref{buildingBlockAlg}. Thus, given that the \textit{Condition 1} holds, the goal reachability of Algorithm \ref{buildingBlockAlg} is guaranteed.
	
Figure 1 illustrates a situation, where, the Condition 1 holds and Figure 2 illustrates a situation, where the Condition 1 is violated.  In Figure 1, $s$ is search frontier for current execution of the while loop of generateLSS with $g(s) = 0, g(x) = \infty$, for each $x \in \{a,b,c,d,e,f,g,h\}$. After $s$ is expanded, for each $x \in \{a,b,c,d,e,f,g,h\}$, we have $g(x)=cost(s,x)$. The next execution $a$ becomes the search frontier (shown in Figure 2). For the state $a$, goal reachability condition is violated, as none of its vacant neighboring states has $g$ value that is greater than $g(a)+cost(a,x')$, for each $x' \in \{h,s,\}$. Thus, the Open list is not updated and the search stops, as it does not have any available move.             


\begin{figure}[ht]
\centering

\vspace{0.5cm}{}
\label{fig:C1R}
\begin{tabular}{|
>{\columncolor[HTML]{FFFFFF}}l |
>{\columncolor[HTML]{FFFFFF}}l |
>{\columncolor[HTML]{FFFFFF}}l |}
\hline
a & b & c \\ \hline
h & \cellcolor[HTML]{036400}s & d \\ \hline
g & f & e \\ \hline
\end{tabular}
\caption{Condition 1 Respected}
\end{figure}

\begin{figure}[ht]
\label{fig:C1V}

\vspace{0.5cm}{}
\centering
\begin{tabular}{|
>{\columncolor[HTML]{333333}}l |l|l|l|}
\hline
 & \cellcolor[HTML]{333333}  & \cellcolor[HTML]{333333}  & \cellcolor[HTML]{333333} \\ \hline
 & \cellcolor[HTML]{036400}a & \cellcolor[HTML]{FFFFFF}b & c                        \\ \hline
 & \cellcolor[HTML]{FFFFFF}h & s & d                        \\ \hline
 & g                         & f                         &  e                        \\ \hline                    
\end{tabular}
\caption{Condition 1 Violated}
\end{figure}


\section{Empirical Evaluation}
In our experiments, we have used pathfinding in video game maps as our test-bed. A game map is represented as 8-connected 2D discrete grid. The grid is composed of cells. In video game maps, cells can be categorized into two types: vacant cell (white) and blocked cell (black). At a given time, an agent can occupy only one cell and in the next time it can move into a neighboring vacant cell. The travel cost associated with the diagonal movement and cardinal movement are $\sqrt{2}$ and 1 respectively. 

We have performed some large-scale experiments with our implementation. Our game maps benchmarks are taken from Moving AI [Sturtevant et. al. 2012].

First, we present the results on our first experiment, where we keep all building block values fixed, except the lookahead depth \texttt{lookahead}, to find out its effect on suboptimality and scrubbing complexity. Then we present experimental results on the simulated evolution of our building block framework for three different settings.

\subsection{Lookahead, Suboptimality and Scrubbing Complexity}
To observe the effect of increasing lookahead depth on suboptimality and scrubbing complexity, we have performed an experiment by keeping all the building fixed except lookahead depth \texttt{lookahead}. For this experiment, we have used 300 random path finding problems.

Figure 3 shows the effect of increasing lookahead depth on suboptimality and scrubbing complexity. In general, the suboptimality decreases with increased lookahead depth. 
	
\begin{figure}[ht] 
	\label{LvSbSc}
	\includegraphics[width=0.5\textwidth]{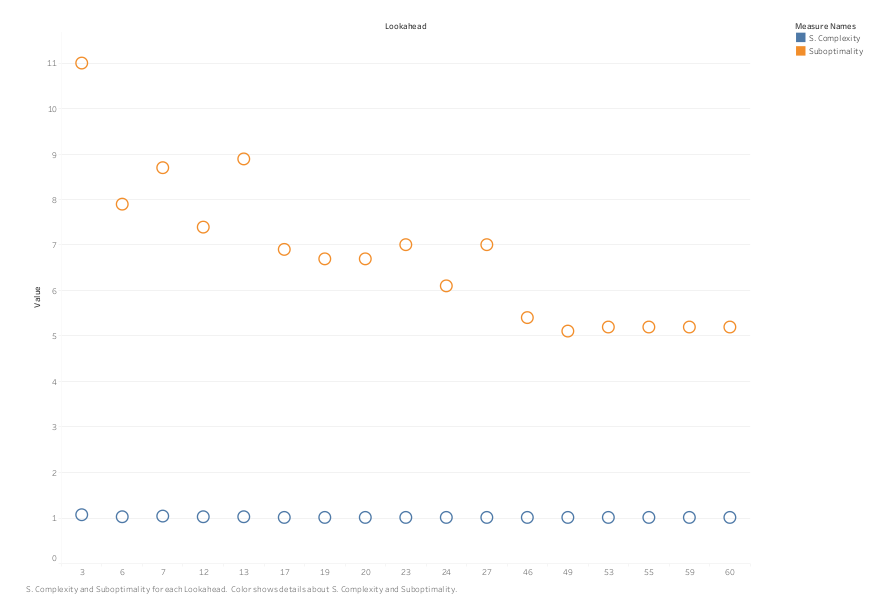} 
	\caption{Lookahead VS Suboptimality, Scrubbing Complexity}
\end{figure}

\subsection{Simulated Evolution}
We have performed simulated evolution of the space of algorithms, which is constituted with our building blocks. Each algorithm in the space of algorithms represents an agent. An agent is represented by its gene, where, a gene is a vector ($w, lop, da$, \texttt{lookahead}, \texttt{lookaheadMethod}). Intuitively, a distinct combination of building block values corresponds to a distinct gene. Table \ref{tab:BBVR} shows the minimum and maximum values allowed for various building blocks of a gene. In a gene, a randomly generated value for building blocks $da,lop$, \texttt{lookahead} and \texttt{lookaheadMethod} are rounded to their nearest integers. We denote a gene by the following the expression: $w$. $lop(c +h)+da+\texttt{lookahead}+\texttt{lookaheadMethod}$.
\begin{table}[t]
	\caption{\small Building Blocks value range}
	\label{tab:BBVR}
	\vspace{0.1cm}
	{\small \begin{center}
		\begin{tabular}{c|c|c}
			\hline
			Building Block           & Min. Value & Max. Value \\
			$w$                      & 1          & 3          \\
			$da$                     & 1          & 2          \\
			$lop$                    & 1          & 2          \\
			\texttt{lookahead}       & 2          & 80         \\
			\texttt{lookaheadMethod} & 1          & 2          
		\end{tabular}
		\end{center}}
\end{table}

We adopted an implementation of a genetic algorithm that is faithful to the Algorithm 4 of \cite{bulitko2016evolving}. The code for this implementation is provided to us by the course instructor.

In our experiment of simulated evolution, we have performed three evolution run. 

In the first run, we have used 16 genes and 10 generations, that is, in each generation of the simulated evolution, we have ran 16 genes. Each agent of each generation
was evaluated on 400 random problems. The suboptimality cutoff is set to 1000, that is, while solving a problem if any agent reaches the travel cost which exceeds 1000 times of optimal travel cost, we just terminate execution for that gene on that problem. The minimum suboptimality achieved by the simulated evolution is 1.74. It was first achieved in generation 4 by the algorithm  $1.1943 · min(c + h)+da+59+A^{*}$. The run took approximately 14 hours.

In the second run, we have used 22 genes and 10 generations, that is, in each generation of the simulated evolution, we have ran 22 genes. Each agent of each generation
was evaluated on 400 random problems. The suboptimality cutoff is set to 1000. The minimum suboptimality achieved by the simulated evolution is 1.96. It was first achieved in generation 10 by the algorithm  $1.2717 · min(c + h)+da+40+A^{*}$. The run took approximately 30 hours.

In the third run, we have used 22 genes and 10 generations, that is, in each generation of the simulated evolution, we have run 22 genes. Each agent of each generation was evaluated on 400 random problems. The suboptimality cutoff is set to 1000. The minimum suboptimality achieved by the simulated evolution is 2.13. it was first achieved in generation 12 by the algorithm  $1.1445 · min(c + h)+da+31+A^{*}$. The run took approximately 32 hours.

Table 2 shows best algorithms determined by our three run of simulated evolution.

\begin{figure}[ht] 
	\label{fig:gen1}
	\includegraphics[width=0.5\textwidth]{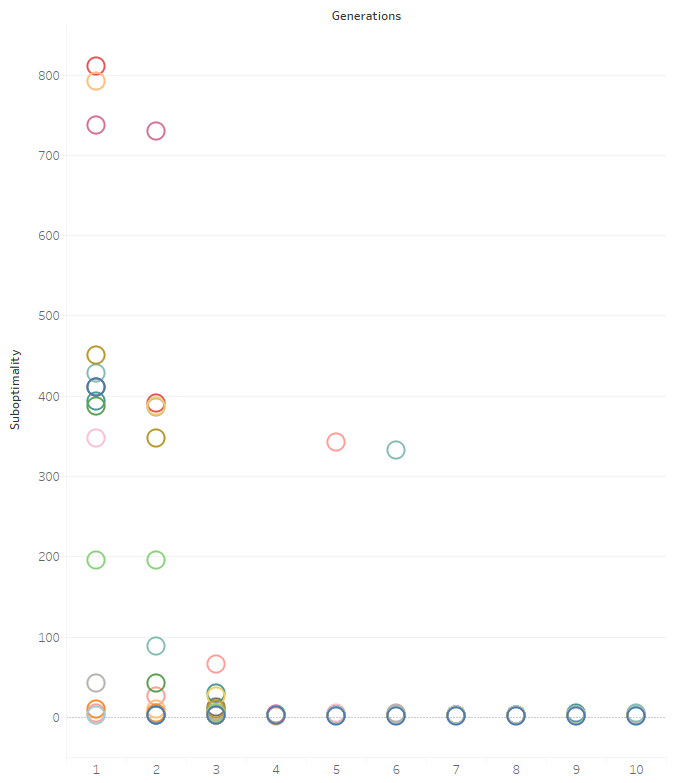} 
	\caption{Change of Suboptimality across generation (First Evolution Run)}
\end{figure} 
	   
\begin{figure}[ht] 
	\label{fig:gen2}
	\includegraphics[width=0.5\textwidth]{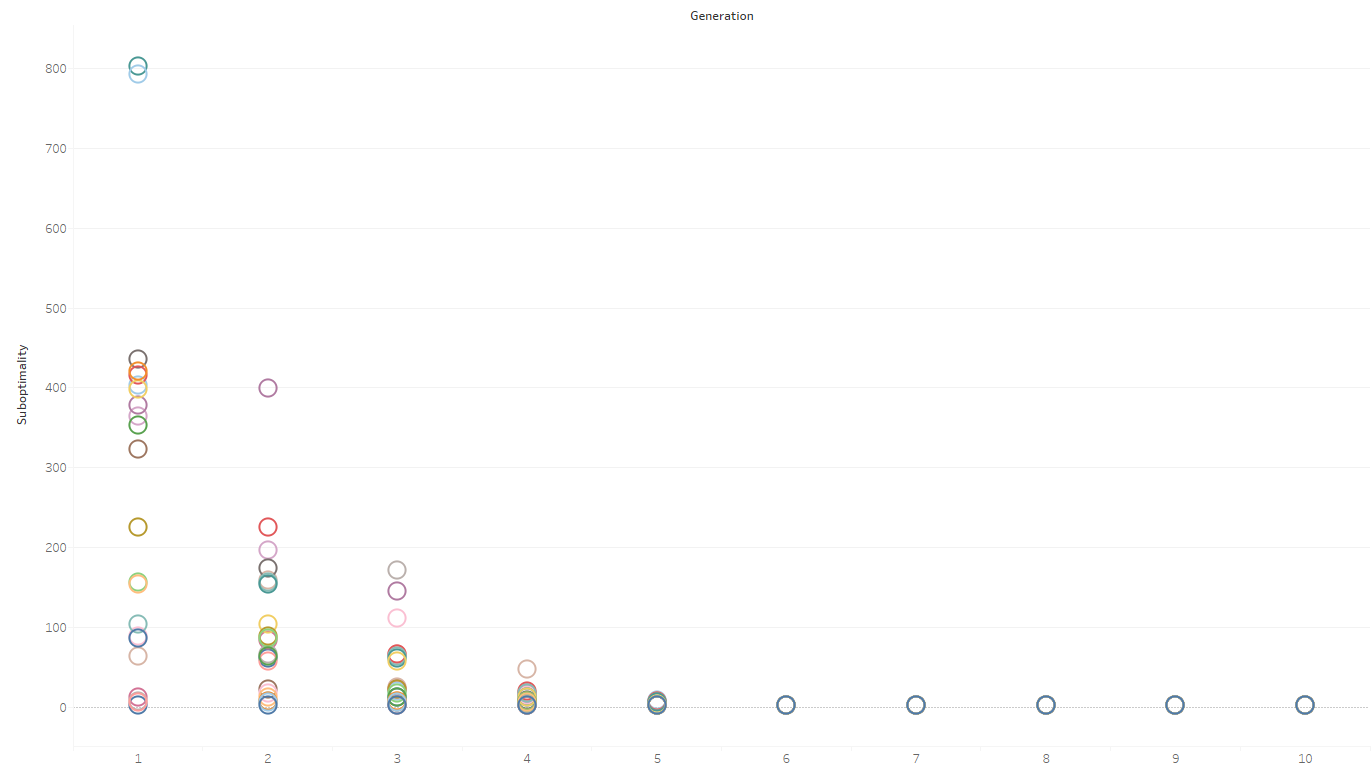} 
	\caption{Change of Suboptimality across generation (Second Evolution Run)}
\end{figure} 
    
\begin{figure}[ht] 
    \label{fig:gen3}
	\includegraphics[width=0.5\textwidth]{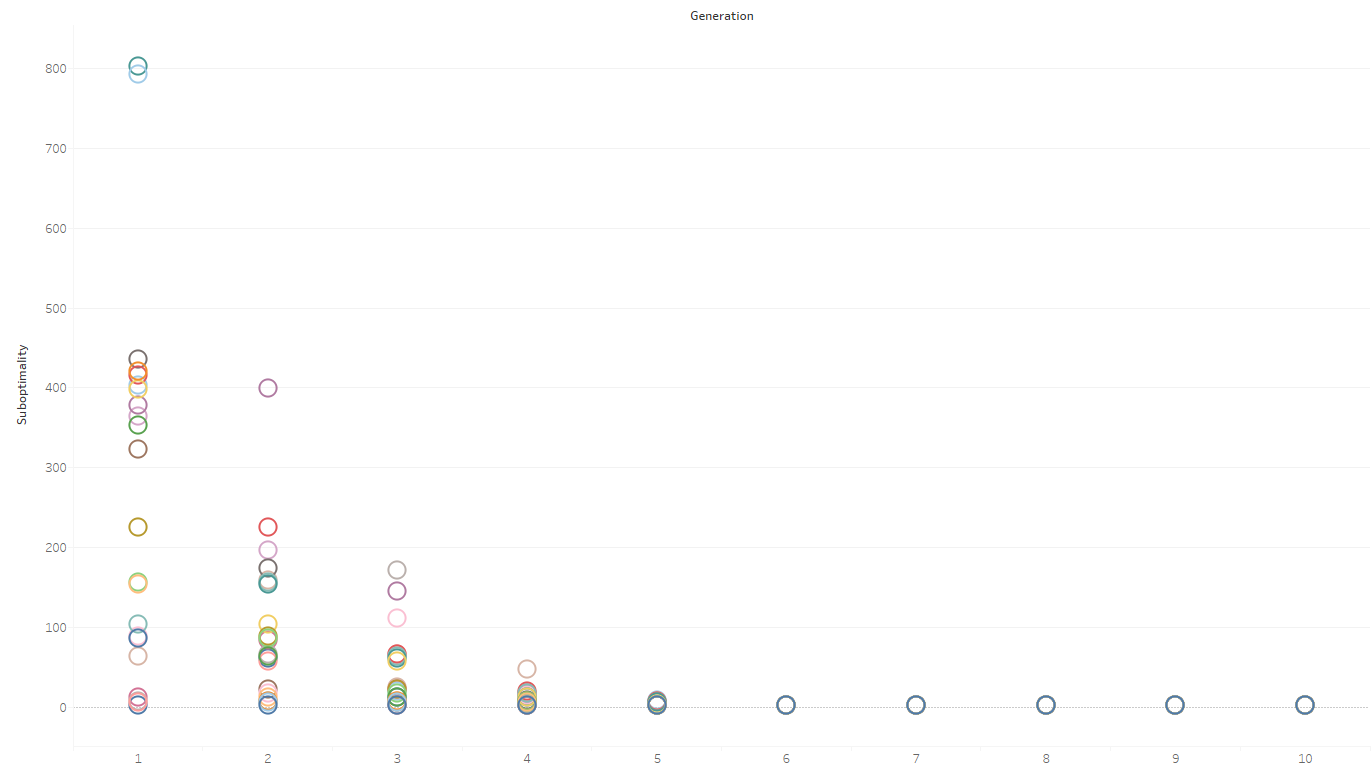} 
	\caption{Change of Suboptimality across generation (Third Evolution Run)}
\end{figure} 
	   
\begin{table}[H]
\label{results}
   \centering
\begin{tabular*}{0.5\textwidth}{c|c|c}
			\hline
			Algorithm                                     & Suboptimality & Scrubbing Complexity \\
			\tiny	$1.1943 \;·\; min(c + h)+da+59+A^{*}$ & \tiny 1.74          & \tiny  1.0019               \\
			\tiny	$1.2714 \;·\; min(c + h)+da+40+A^{*}$ & \tiny 1.96          & \tiny 1.0026               \\
			\tiny	$1.1445 \;·\; min(c + h)+da+31+A^{*}$ & \tiny 2.13          & \tiny 1.0032               
\end{tabular*}
\end{table}

	Figure 4, Figure 5 and Figure 6 shows the convergence of suboptimality value with the increase of generations for the first, second and third evolution run respectively. Each circle with a distinct color represents a gene. For the first, second and third evolutionary run, the evolutionary process starts to converge from the 4th, 5th and 6th generation respectively.  

	\section{Discussion}
	\subsection{Discussion on Lookahead VS Suboptimality}
	In our first experiment, keeping other building block value fixed, we only allowed the lookahead depth \texttt{lookahead} to vary. From the Figure 3, we see that lookahead depth of 3 gives us the highest suboptimality and lookahead depth of 49 gives us the lowest suboptimality. In Figure 3, a) in general, suboptimality decreases with the increases of lookahead depth. b) There are some cases, where with the increase of lookahead depth, suboptimality increases or stays the same. The reduction of scrubbing complexity is also observed with the increase of lookahead depth.
			
	Roughly speaking, this experimental result is consistent with the experimental results of \cite{sven09lssLRTA}. As shown in Table 6 of \cite{sven09lssLRTA}, for Grid with random obstacle benchmark, with the increase of lookahead depth, trajectory length (i.e., suboptimality) generally decreases. However, there are some exceptions observed in Table 6, where trajectory length increases or stays the same with increase of lookahead depth. For example, for lookahead depth from 29 to 49 (in Table 6 of \cite{sven09lssLRTA}), the incurred trajectory lengths do not consistently increase.
			
	\subsection{Discussion on Simulated Evolution}
	In our simulated evolution experiment, for all three of the evolution runs, suboptimality value converges nicely. For smaller space of algorithms, the simulated evolution reaches convergence earlier, than with the larger space of algorithms. One possible explanation of this scenario is as following: the Simulated evolution explores more with larger space of algorithms than it does with smaller space of algorithms. This results in delay in convergence for the larger space of algorithms.
				
	The best three algorithms from three of our simulation runs achieves low suboptimality and scrubbing complexity. The suboptimality and scrubbing complexity achieved by these three algorithms are close. With smaller space of algorithms, the simulated evolution produces slightly better suboptimality than with the larger space of algorithms. We have observed that for the first few generations (before convergence), for smaller space of algorithms, the simulated evolution process produces genes that have uniformly distributed values for all the building block. On the other hand, for larger space of algorithms, for the first few generations, the value distributions for the building blocks with larger range of values, such as \texttt{lookahead}, are less uniformly distributed. This causes emergence of genes (in the first few generations for larger space of algorithms), in which \texttt{lookahead} values tend to concentrate mostly into the lower half of the allowable range ([1,80]).  Thus, for larger space of algorithms, though more exploration is performed with other building blocks, less exploration is performed with \texttt{lookahead}. Consequently, for larger space of algorithms, the simulated evolution process converges to a less optimal gene than it does for smaller space of algorithms.
				
	Overall, these results show the power of using LSS-LRTA* style lookahead method in our building block. Deeper exploration using $A^{*}$, combined with depression avoidance and small weighted learning appears to be really effective. 
				
	Though Goal Achievement Time (GAT) is not formally reported here, in general, we observed that, with genes with larger lookahead depth, GAT also reduces, compared to genes with smaller lookahead. Genes with larger lookahead depth follows better optimal path, yields less scrubbing and thus GAT reduces.
				
	\section{Conclusions and Future Work}
    	In this paper, we have extended the building block framework of \cite{bulitko2016evolving}, by adding  LSS-LRTA* style lookahead based real-time heuristics search algorithms. To accomplished this, first, by adopting the algorithms from \cite{sven09lssLRTA} and then by accommodating other building blocks from \cite{bulitko2016evolving}. We have presented our theoretical analysis on termination and goal reachability of our building block based real-time heuristics search algorithm. Then, we implemented these algorithms in MATLAB. Next, we perform experiments on path finding problems from the video game domain. The first experiment shows that in general, with increased lookahead depth suboptimality decreases. This reassures the observation of \cite{sven09lssLRTA}. In the second experiment, we have performed simulated evolution of the extended building framework. This experiment reveals that the best algorithms from our extended space of algorithms achieve low suboptimality and scrubbing complexity. This reassures the power of lookahead based real-time heuristics search algorithms.
        
	As future work, we plan to investigate more to solve the implementation issues that we encountered. Adding other building blocks, such as, beam width $b$, \texttt{expendable}  and \texttt{backtrack} in our building block framework and performing more experiments on the extended space of algorithms will be an interesting extension of our paper. Currently, as fitness function, our simulated evolution uses only suboptimality of the algorithms. Performing simulated evolution with a linear combination of metrics such suboptimality, scrubbing complexity and goal achievement time is another interesting direction. In the future, we intend to use other benchmarks, such as, maze, to perform simulated evolution with our extended space of algorithms.

	\color{black}
		
	\section*{Acknowledgement}
	This work was done for a graduate course project at the Computing Science department of the University of Alberta. We thank our instructor Dr. Vadim Bulitko for providing the initial code base for the project, in addition to his feed-backs and suggestions throughout the course project.
  	\bibliographystyle{aaai}

	\bibliography{projectReport}
\end{document}